\documentclass{article}
\usepackage{ijcai25}

\newcommand{\citet}[1]{\citeauthor{#1}~(\citeyear{#1})}

\usepackage{times}
\usepackage{soul}
\usepackage{url}
\usepackage[hidelinks]{hyperref}
\usepackage[utf8]{inputenc}
\usepackage[small]{caption}
\usepackage{graphicx}
\usepackage{amsmath}
\usepackage{amsthm}
\usepackage{booktabs}
\usepackage{multirow}
\usepackage{ragged2e}

\usepackage{algorithm}
\usepackage{algorithmic}
\usepackage[switch]{lineno}
\usepackage{setspace}

\usepackage[normalem]{ulem}
\useunder{\uline}{\ul}{}

\usepackage{float,epsfig,subfig,color}
\usepackage{enumitem}
\usepackage{algorithmic,algorithm}
\usepackage{comment}

\usepackage{amsmath,amsfonts,bm}

\def\eqref#1{equation~(\ref{#1})}
\def\Eqref#1{Eq.~(\ref{#1})}

\def\1{\bm{1}}

\def\rd{{\mathrm{d}}}

\def\vh{{\bm{h}}}

\def\vk{{\bm{k}}}

\def\vx{{\bm{x}}}

\def\vz{{\bm{z}}}

\DeclareMathAlphabet{\mathsfit}{\encodingdefault}{\sfdefault}{m}{sl}
\SetMathAlphabet{\mathsfit}{bold}{\encodingdefault}{\sfdefault}{bx}{n}

\def\gG{{\mathcal{G}}}

\newcommand{\logsig}{\mathrm{LogSig}}

\usepackage{array}
\newcommand{\PreserveBackslash}[1]{\let\temp=\\#1\let\\=\temp}
\newcolumntype{C}[1]{>{\PreserveBackslash\centering}p{#1}}
\newcolumntype{R}[1]{>{\PreserveBackslash\raggedleft}p{#1}}
\newcolumntype{L}[1]{>{\PreserveBackslash\raggedright}p{#1}}
\newcolumntype{P}[1]{>{\raggedright\arraybackslash}m{#1}}

\newcommand{\std}[1]{\tiny{$\pm$#1}}

\title{Comprehensive Review of Neural Differential Equations for Time Series Analysis}

\author{
YongKyung Oh$^1$\and \;
Seungsu Kam$^2$\and \;
Jonghun Lee$^3$\and \;
Dong-Young Lim$^{2,3}$\and \\ % \;
Sungil Kim$^{2,3}$\footnotemark[1]\And % \;  
Alex A. T. Bui$^1$\thanks{Corresponding authors}\\
\affiliations
$^1$Medical \& Imaging Informatics (MII) Group, University of California, Los Angeles (UCLA) \\
$^2$Department of Industrial Engineering, Ulsan National Institute of Science and Technology (UNIST) \\
$^3$Artificial Intelligence Graduate School, Ulsan National Institute of Science and Technology (UNIST) \\
\emails
yongkyungoh@mednet.ucla.edu, \;
\{lewki83, jh.lee, dlim, sungil.kim\}@unist.ac.kr, \;
buia@mii.ucla.edu
}
\begin{document}

\maketitle

\begin{abstract}
    Time series modeling and analysis have become critical in various domains. Conventional methods such as RNNs and Transformers, while effective for discrete-time and regularly sampled data, face significant challenges in capturing the continuous dynamics and irregular sampling patterns inherent in real-world scenarios. Neural Differential Equations (NDEs) represent a paradigm shift by combining the flexibility of neural networks with the mathematical rigor of differential equations. This paper presents a comprehensive review of NDE-based methods for time series analysis, including neural ordinary differential equations, neural controlled differential equations, and neural stochastic differential equations. We provide a detailed discussion of their mathematical formulations, numerical methods, and applications, highlighting their ability to model continuous-time dynamics. Furthermore, we address key challenges and future research directions. This survey serves as a foundation for researchers and practitioners seeking to leverage NDEs for advanced time series analysis.
\end{abstract}

%-----------------------------------------------------------------------
\section{Introduction}\label{sec:Introduction}
The exponential growth of time series data across diverse domains has necessitated more sophisticated analytical approaches, because of the complex nonlinear nature, irregular sampling, missing values, and continuous latent dynamics.  
Traditional approaches~\cite{durbin2012time,box2015time} assume regular sampling and often linear relationships, limiting their applicability to real-world scenarios. 
While deep learning methods, including Recurrent Neural Network (RNN)~\cite{rumelhart1986learning,medsker1999recurrent}, Long Short-term Memory (LSTM)~\cite{hochreiter1997long}, Gated Recurrent Unit (GRU)~\cite{chung2014empirical}, and Transformer~\cite{vaswani2017attention} have shown promise in capturing nonlinearity, they remain constrained by discrete-time formulations \cite{che2018recurrent,sun2020review,weerakody2021review}

\emph{Neural Differential Equations} (NDEs) emerged as a paradigm shift, offering a principled framework for continuous-time modeling with neural networks. The seminal work on Neural Ordinary Differential Equations (NODEs)~\cite{chen2018neural} introduced continuous-time hidden state evolution, spawning numerous advances including Neural Controlled Differential Equations (NCDEs)~\cite{kidger2020neural} incorporating external control paths, and Neural Stochastic Differential Equations (NSDEs)~\cite{han2017deep,tzen2019neural,li2020scalable,oh2024stable} modeling uncertainty through stochasticity.

In this survey, we provide a comprehensive review of NDE methods in time series analysis, synthesizing advancements across several interconnected areas. We begin by exploring the base model families, including NODEs, NCDEs, and NSDEs, along with their extensions and variations. Building on this, we analyze the theoretical insights that underpin these models. 
We also discuss the implementation of NDE methods and analyze applications. This survey aims to provide a structured synthesis for researchers and practitioners, fostering a deeper understanding of NDEs and their potential in tackling complex time series problems.

%-----------------------------------------------------------------------
\section{Preliminaries}\label{sec:Preliminaries}
\begin{figure*}[!htb]
\centering
\captionsetup{skip=5pt}
\captionsetup[subfigure]{justification=centering, skip=5pt}
\subfloat[State of piecewise interpolations]{
  \includegraphics[clip,width=0.32\linewidth]{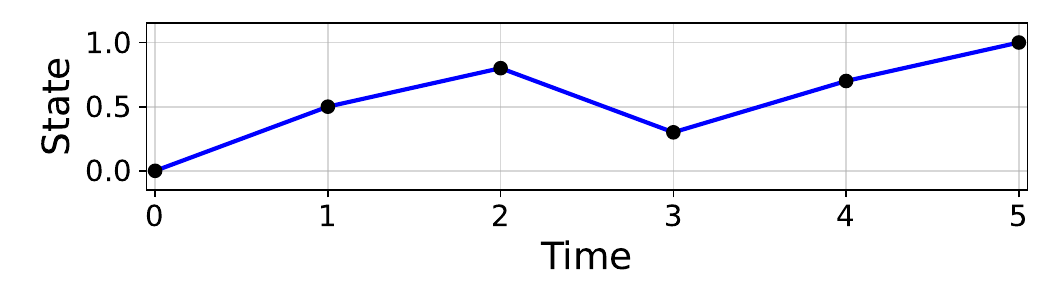}} \hfil
\subfloat[State of traditional RNNs]{
  \includegraphics[clip,width=0.32\linewidth]{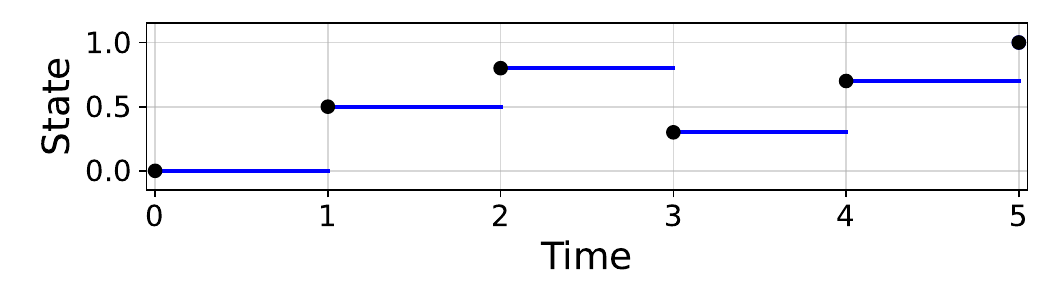}} \hfil
\subfloat[State of NDE-based methods]{
  \includegraphics[clip,width=0.32\linewidth]{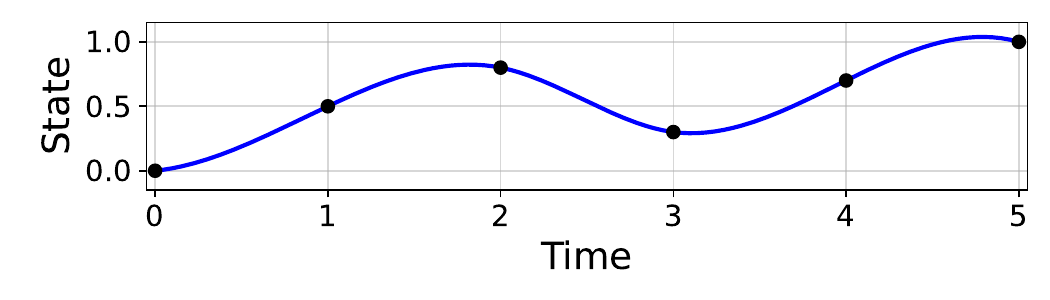}} \\ [0pt]
\subfloat[Gradient of piecewise interpolations]{
  \includegraphics[clip,width=0.32\linewidth]{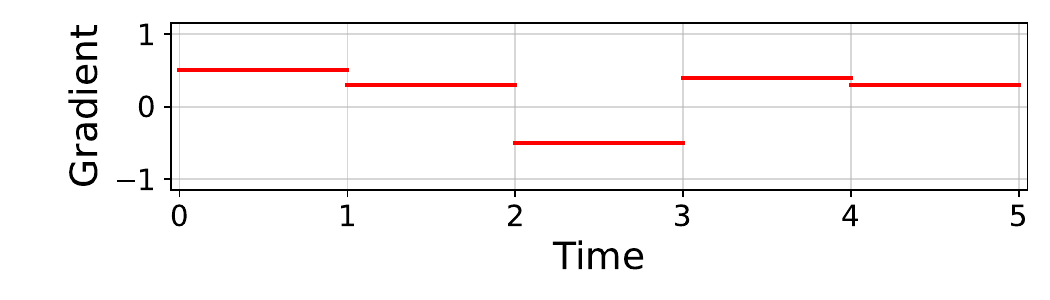}} \hfil
\subfloat[Gradient of traditional RNNs]{
  \includegraphics[clip,width=0.32\linewidth]{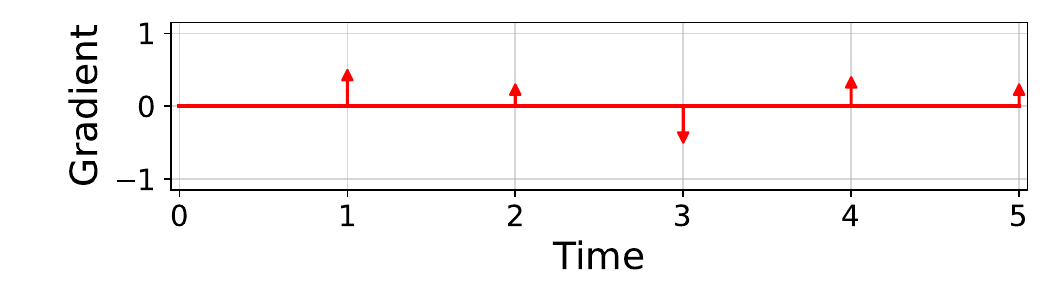}} \hfil
\subfloat[Gradient of NDE-based methods]{
  \includegraphics[clip,width=0.32\linewidth]{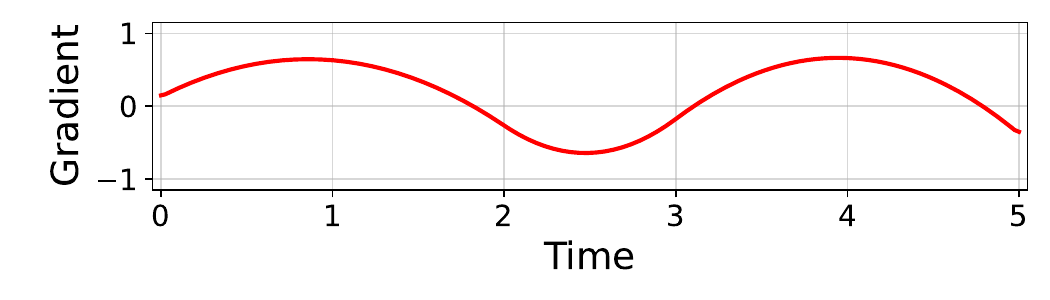}}
\caption{Conceptual comparison of time series modeling. (a) Piecewise interpolation provides continuity but lacks dynamic modeling capabilities. (b) Traditional RNNs operate on discrete steps, encountering limitations with continuous dynamics, irregularly sampled data, and memory efficiency for long sequences. (c) NDE-based methods, such as NODEs, NCDEs, and NSDEs, learn continuous dynamics, inherently accommodate irregular sampling, and offer potential memory advantages, when employing techniques like the adjoint method.}
\label{fig:ex}
\end{figure*}

Time series modeling seeks to represent sequential data \(\vx = (x_0, x_1, \ldots, x_n)\), where each \(x_i \in \mathbb{R}^{d_x}\), as a continuous latent process \(\vz(t) \in \mathbb{R}^{d_z}\) over a time domain \([0,T]\). 
Figure~\ref{fig:ex} illustrates the conceptual differences between conventional time series modeling and NDE-based methods. While interpolation methods simply fit curves through observations (black dots) and RNNs (often using zero-order hold) operate in discrete time steps, NDEs aim to learn the underlying continuous dynamics, $\rd \vx(t)/\rd t$ (or $\rd \vz(t)/\rd t$), parameterized by a neural network~\cite{chen2018neural}. This parameterization provides NDEs with a flexible framework for handling challenges common in real-world time series, such as irregular sampling, missing observations, and capturing long-horizon dependencies. Furthermore, NDEs can offer greater memory efficiency compared to standard discrete-time architectures~\cite{chen2018neural,kidger2020neural,rubanova2019latent,oh2024stable}.

RNNs, including LSTMs~\cite{hochreiter1997long} and GRUs~\cite{chung2014empirical}, operate in discrete time, updating hidden states sequentially based on observed inputs. Transformer-based models~\cite{vaswani2017attention} offer an alternative by leveraging self-attention mechanisms, though they typically rely on fixed positional encodings and do not inherently capture continuous-time dynamics.

Various modifications have been proposed to introduce a notion of time awareness in RNNs~\cite{che2018recurrent,rajkomar2018scalable}.
One approach adjusts updates for time intervals, while another uses decay mechanisms to approximate continuous evolution.
However, these heuristics do not always generalize across different settings and, in some cases, have been found to perform comparably to standard RNNs~\cite{mozer2017discrete,rubanova2019latent}.

\subsection{Neural Ordinary Differential Equations}\label{sec:nodes}
NODEs~\cite{chen2018neural} model the  latent state \(\vz(t)\) as:
\begin{equation}\label{eq:neural_ode}
    \vz(t) = \vz(0) + \int_0^t f\bigl(s, \vz(s); \theta_f\bigr) \, \mathrm{d}s,
\end{equation}
where \(\vz(0) = h(\vx; \theta_h)\), \(h: \mathbb{R}^{d_x} \to \mathbb{R}^{d_z}\) is a neural network parameterized by \(\theta_h\), and \(f\bigl(t, \vz(t); \theta_f\bigr)\) approximates \(\frac{\mathrm{d}\vz(t)}{\mathrm{d}t}\). The vector field \(f\) is typically implemented using multi-layer perceptrons or more sophisticated architectures. 
One can view these approaches as an infinite-layer generalization of residual networks, allowing integration over time rather than discrete stacking of layers \cite{chen2018neural,rubanova2019latent,dupont2019augmented,kidger2020neural}.

\subsection{Neural Controlled Differential Equations}\label{sec:ncdes}
NCDEs~\cite{kidger2020neural} extend NODEs by incorporating a control path \(X(t)\) for updating state over time:
\begin{equation}\label{eq:neural_cde}
    \vz(t) = \vz(0) + \int_0^t f\bigl(s, \vz(s); \theta_f\bigr) \, \mathrm{d}X(s),
\end{equation}
where the integral is interpreted as a Riemann–Stieltjes integral, allowing for discontinuous paths. 
Piecewise-smooth control paths \(X(t)\) are typically constructed using natural cubic splines~\cite{kidger2020neural} or Hermite cubic splines~\cite{morrill2021neural}. Furthermore, \citet{morrill2021rough} extend Neural CDEs by incorporating rough path theory for the generalized formulation.

\subsection{Neural Stochastic Differential Equations}\label{sec:nsdes}
NSDEs~\cite{han2017deep,tzen2019neural,li2020scalable} incorporate stochasticity through:
\begin{align}\label{eq:neural_sde}
    % \vz(t) = \vz(0) + \int_0^t f\bigl(s, \vz(s); \theta_f\bigr) \, \mathrm{d}s + \int_0^t g\bigl(s, \vz(s); \theta_g\bigr) \, \mathrm{d}W(s),
    \vz(t) = \vz(0) 
    & + \int_0^t f\bigl(s, \vz(s); \theta_f\bigr) \, \mathrm{d}s \nonumber \\
    & + \int_0^t g\bigl(s, \vz(s); \theta_g\bigr) \, \mathrm{d}W(s), 
\end{align}
where \(W(t)\) is a Wiener process (or Brownian motion), \(f\bigl(\cdot; \theta_f\bigr)\) is the drift function, and \(g\bigl(\cdot; \theta_g\bigr)\) is the diffusion function. The stochastic integral follows It\^o or Stratonovich interpretations. 
While NODEs describe deterministic evolution through ordinary differential equations, NSDEs model uncertainty and noise through Brownian motion terms \cite{jia2019neural,liu2019neural}. This extension enables more robust modeling of real-world phenomena where randomness plays a crucial role \cite{kidger2021neuralsde,kidger2021efficient,oh2024stable}, each offering different approaches to balance stability, expressivity, and computational tractability.

%-----------------------------------------------------------------------
\section{Theoretical Considerations}\label{sec:theoretical}
% The theoretical foundations of NDEs are essential for understanding their ability to model complex temporal dynamics. Their connection to dynamical systems explains how they capture continuous-time evolution, while universal approximation properties justify their flexibility in learning intricate patterns. Additionally, stability and convergence analysis ensures robustness, making NDEs a reliable framework. 

\subsection{Dynamical Systems Perspective}\label{sec:theoretical_dynamical}
NDEs generalize the notion of discrete layers (as in RNNs or Transformers) to a continuous-time vector field. According to \Eqref{eq:neural_ode}, the latent representation $\vz(t)$ evolves:
\begin{equation}\label{eq:node}
    \frac{\mathrm{d}\vz(t)}{\mathrm{d}t} = f\bigl(t, \vz(t); \theta_f\bigr),
\end{equation}
while in NCDEs, the evolution is driven by a control path: % $X(t)$:
\begin{equation}\label{eq:ncde}
    \mathrm{d}\vz(t) = f\bigl(t, \vz(t); \theta_f\bigr)\,\mathrm{d}X(t).
\end{equation}
Here, $X(t)$ is typically a spline or piecewise interpolation of observed input data $\vx$, as explained in \Eqref{eq:neural_cde}.

Stochastic extensions, such as NSDEs, incorporate Brownian motion $W(t)$ or jump processes to model noise or uncertainty in the underlying dynamics, as shown in \Eqref{eq:neural_sde}: 
\begin{equation}\label{eq:nsde}
    \mathrm{d}\vz(t) 
    = f\bigl(t, \vz(t);\theta_f\bigr)\,\mathrm{d}t 
    + g\bigl(t, \vz(t);\theta_g\bigr)\,\mathrm{d}W(t),    
\end{equation}
where \(W(t)\) is a Wiener process.
Formally, these methods place the latent trajectory $\vz(t)$ in a phase space:
\begin{equation}\label{eq:initial}
    \vz\colon [0, T] \to \mathbb{R}^{d_z}, 
    \quad \vz(0) = h(\vx;\theta_h),
\end{equation}
where $h: \mathbb{R}^{d_x} \rightarrow \mathbb{R}^{d_z}$ maps raw inputs $\vx$ to the initial state. 
By solving the equations above, one obtains a continuous phase-space trajectory that can be analyzed through classical dynamical systems tools.
This continuous viewpoint also enables seamless continuous-time modeling.

\paragraph{Invertible Neural Networks and Flow Models.}
In parallel with NDEs, explicit approaches like \emph{Neural Flows} have also been proposed \cite{lu2018beyond,sonoda2019transport,grathwohl2018scalable,massaroli2020stable,bilovs2021neural}. Rather than parameterizing the rate of change $\mathrm{d}\vz(t)/\mathrm{d}t$, Neural Flows directly model the solution map: 
\begin{equation}\label{eq:neural_flow}
\vz(t) = \mathcal{F}\bigl(t,\vz(0);\theta_{\mathcal{F}}\bigr),
\end{equation}
where $\vz(0) = h(\vx;\theta_h)$ is the initial condition. This explicit representation bypasses the need for numerical ODE solvers, potentially offering faster computations or enhanced stability via invertibility constraints \cite{kobyzev2020normalizing,papamakarios2021normalizing}. Such architectures align with normalizing-flow models, which are invertible by construction and allow for efficient likelihood evaluation.

\paragraph{Contrasting Implicit and Explicit Methods.}
The distinction between NODEs/NCDEs (which define $\mathrm{d}\vz/\mathrm{d}t$ and rely on solvers) and Neural Flows (which specify $\vz(t)$ directly) parallels the classic implicit versus explicit solution dichotomy in differential equations. Implicit, solver-based methods (like NODEs, NCDEs, NSDEs) can handle irregular inputs through fixed or adaptive step sizes and are more general in capturing unknown or complex dynamics. Explicit approaches, like Neural Flows~\cite{bilovs2021neural}, may yield simpler training pipelines for stable transformations, yet can suffer from constraints on network forms and from instabilities if the sought-after closed-form solution is nonexistent or approximate \cite{oh2025dualdynamics}. 

In this survey, we limit our scope to NDEs that require implicit solvers, excluding flow-based models and other explicit parameterizations of continuous-time dynamics.

\subsection{Universal Approximation Properties}\label{sec:theoretical_approximation}

\paragraph{NODEs.}
\citet{chen2018neural} established that continuous-time models can be viewed as an infinite-depth limit of residual networks, enabling them to approximate diffeomorphic transformations under mild regularity conditions, as represented in \Eqref{eq:node}.
However, autonomous ODE flows may limit expressivity by constraining trajectories to remain diffeomorphic transformations of initial states, thus prohibiting intersecting paths in latent space.
\citet{dupont2019augmented} addressed these limitations through augmented ODEs, which append an auxiliary variable \(\bm{a}(t)\) to the state:
\begin{equation}
    \frac{\mathrm{d}}{\mathrm{d}t} 
    \begin{pmatrix}
    \vz(t) \\
    \bm{a}(t)
    \end{pmatrix}
    = f\Bigl(t, \vz(t), \bm{a}(t);\theta_f\Bigr).
\end{equation}
Augmenting the latent dimension effectively bypasses strict diffeomorphic constraints, expanding the range of representable data manifolds. Empirically, this technique improves reconstruction and expressivity in tasks where standard Neural ODEs struggle due to their trajectory overlap constraint.

\paragraph{NCDEs.}
While NODEs rely solely on a learned vector field and initial value, \citet{kidger2020neural} showed that Neural CDEs embed a control path \(X(t)\) into the model as explained in \Eqref{eq:ncde}:
By allowing data updates to arrive as increments \(\mathrm{d}X(t)\) at arbitrary times, NCDEs implement a form of continuous recursion, demonstrating universal approximation for continuous paths under sufficiently rich control signals. Extensions to rough paths \cite{morrill2021rough} further expand coverage to long, irregular time series, handling signals too irregular for traditional ODE integrators.

\paragraph{NSDEs.}
Stochastic extensions achieve universal approximation for broad classes of continuous-time random processes by learning drift and diffusion terms. \citet{han2017deep} and \citet{tzen2019neural} formalized how parameterizing \(f\) and \(g\) via neural networks lets these models approximate various stochastic phenomena, such as Geometric Brownian motion or Ornstein–Uhlenbeck processes. Subsequent works generalize to jump processes and specialized SDE families \cite{jia2019neural,li2020scalable,oh2024stable,zhang2024neural}, underscoring the versatility of NSDEs in capturing noise-driven dynamics.

%%%
\subsection{Existence and Uniqueness of Solutions}\label{sec:theoretical:theoretical_existence}

\paragraph{Deterministic NDEs.}
For NODEs as explained in \Eqref{eq:node}, classic results such as the Picard–Lindelöf theorem guarantees that a unique solution $\vz(t)$ exists if the vector field $f$ is Lipschitz continuous in $\vz$. Formally, if 
\begin{equation}
    \|f\bigl(t, \vz(t_1); \theta_f\bigr) - f\bigl(t, \vz(t_2); \theta_f\bigr)\| \,\le\, L\,\|\vz(t_1) - \vz(t_2)\|
\end{equation}
for some constant $L>0$, then the integral equation describing $\vz(t)$ has a unique solution starting from $\vz(0)=h(\vx)$ \cite{chen2018neural}. In practice, spectral normalization or weight clipping can enforce such Lipschitz constraints, improving both existence properties and training stability \cite{massaroli2020dissecting,rackauckas2020universal}.

In NCDEs as shown in \Eqref{eq:ncde}, existence and uniqueness in this setting hinge on Lipschitz-like conditions for $f$ and on the path $X(t)$ possessing sufficient regularity (e.g., bounded variation or spline-based interpolation) \cite{kidger2020neural}. If $X(t)$ is discontinuous or highly irregular, rough path theory provides a generalized framework under which well-posedness can still be established \cite{morrill2021neural}. 

\paragraph{Stochastic NDEs.}
NSDEs embed stochasticity via \Eqref{eq:nsde}, existence and uniqueness then require both $f$ and $g$ to satisfy Lipschitz and linear-growth conditions in $\vz$ \cite{tzen2019neural,oh2024stable}. Common conditions for the existence and uniqueness of $\vz$ are Lipschitz continuity and linear growth conditions for $f$ and $g$ such that if there exist constants $L$ and $C$ such that 
\begin{equation}
\begin{aligned}
    &\|f\bigl(\vz(t_1);\theta_f\bigr)-f\bigl(\vz(t_2);\theta_f\bigr)\| \\
    &\quad + \|g\bigl(\vz(t_1);\theta_g\bigr)-g\bigl(\vz(t_2);\theta_g\bigr)\| 
    \,\le\, L\|\vz(t_1)-\vz(t_2)\|
\end{aligned}
\end{equation}
and 
\begin{equation}
    \|f\bigl(\vz;\theta_f\bigr)\| + \|g\bigl(\vz;\theta_g\bigr)\| 
    \,\le\, C\,(1+\|\vz\|),    
\end{equation}
then a unique strong solution $\vz(t)$ exists for all finite $t$. Under these conditions, finite-time blow-up is prevented, ensuring the SDE is well-posed \cite{jia2019neural,li2020scalable}. 
Langevin-type SDEs, linear-noise SDEs, and geometric SDEs—variants of Neural SDEs—are also designed to satisfy these conditions, thereby ensuring the existence and uniqueness of their solutions \cite{oh2024stable}.

%%%
\subsection{Stability and Convergence}\label{sec:theoretical:theoretical_stability}

\paragraph{Deterministic Stability.}
NODEs and NCDEs achieve stability through Lipschitz constraints on vector fields \cite{haber2017stable,chen2018neural,kidger2020neural}. For NODEs, spectral norm conditions on network weights keep trajectories bounded \cite{kidger2021hey}. NCDEs maintain stability by bounding \(\mathrm{d}X/\mathrm{d}t\) \cite{morrill2021neural}.

\paragraph{Stochastic Stability.}
NSDEs follow dynamics \(\mathrm{d}\vz(t) = f\bigl(\vz, t; \theta_f\bigr)\,\mathrm{d}t + g\bigl(\vz, t; \theta_g\bigr)\,\mathrm{d}W(t)\), where stability depends on drift \(f\) and diffusion \(g\) terms \cite{tzen2019neural,kidger2021neuralsde}. 
\citet{oh2024stable} shows the stochastic stability of the proposed Neural SDEs (Langevin-type SDEs, linear-noise SDEs, and geometric SDEs) under suitable regularity conditions and relates the findings to their robustness against distribution shifts.

\paragraph{Optimal Transport and Convergence.}
The convergence of continuous-time methods can be analyzed through optimal transport theory, with the Wasserstein metric offering a principled measure of stability and generalization \cite{villani2009optimal,peyre2019computational}.
Formally, $(\Omega,\mathcal{F},P)$ specifies a probability space for random variables in $L^p(\Omega)$ whenever stochastic components arise. 
For an \(\mathbb{R}^d\)-valued random variable \(X\), its law \(\mathcal{L}(X)\) belongs to \(\mathcal{P}(\mathbb{R}^d)\). The Wasserstein distance of order \(p\) measures how far two distributions \(\mu,\nu\in\mathcal{P}(\mathbb{R}^d)\) are in terms of the minimum cost coupling, such as:
\begin{equation}
    \mathcal{W}_p(\mu, \nu) = \inf_{\Pi \in \mathcal{C}(\mu, \nu)} \Bigl(\int_{\mathbb{R}^d}\!\!\int_{\mathbb{R}^d} |x - x'|^p \,\Pi(\mathrm{d}x, \mathrm{d}x') \Bigr)^{1/p}
\end{equation}
Here, \(\mathcal{C}(\mu, \nu)\) denotes the set of couplings whose marginals match \(\mu\) and \(\nu\). 
In the context of NDE models, particularly in stochastic or high-dimensional regimes, this metric provides a valuable tool for analyzing model robustness, generalization, and convergence properties~\cite{kidger2021on,ruiz2023neural,oh2024stable}.

\paragraph{Solver Convergence.}
NDEs rely on numerical integrators such as Euler, Runge–Kutta, and Dormand–Prince for deterministic models \cite{chen2018neural,rubanova2019latent} and Euler–Maruyama, Milstein, or Reversible Heun for stochastic models \cite{kidger2021efficient,tzen2019neural}. On the other hand, systems that are stiff, exhibit jumps, or involve high-dimensional dynamics often necessitate specialized implicit solvers to ensure stability and computational efficiency \cite{jia2019neural,rackauckas2020universal,kim2021stiff}.

%-----------------------------------------------------------------------
\section{Practical Implementation}\label{sec:practical_implementation}
This section expands on the theoretical foundations from Section~\ref{sec:theoretical}, focusing on the practical aspects of training, regularization, and deployment of NDEs. While discrete-time models are widely studied, continuous-time models remain underexplored despite their advantages. 

\subsection{Optimization of NDE-based Models}
\paragraph{Adjoint Method.}
NDEs differ significantly from discrete methods in their reliance on numerical integration. During training, gradients with respect to model parameters must be backpropagated through an ODE, CDE, or SDE solver, which raises memory and stability challenges. Traditional backpropagation is conceptually straightforward but stores all intermediate states, leading to large memory usage in long sequences or high-dimensional latents. 

The adjoint sensitivity method, introduced by \citet{chen2018neural}, is a pivotal technique for efficiently computing gradients in NDEs by solving an adjoint equation backward in time. This method addresses the challenge of high memory consumption in gradient-based optimization by reconstructing forward states on demand, reducing the memory complexity from \(O(N)\), where \(N\) is the sequence length, to approximately \(O(1)\). The adjoint state \(\lambda(t)\) evolves according to a differential equation, 
\begin{equation}
    \frac{\mathrm{d}\lambda(t)}{\mathrm{d}t} = -\lambda(t)^\top \partial f / \partial \bm{z}    
\end{equation}
where \(\lambda(T) = \partial \mathcal{L}/\partial \bm{z}(T)\) initializes the backward integration. This approach enables the computation of gradients via 
\begin{equation}
    \mathrm{d}\mathcal{L}/\mathrm{d}\theta = \int_0^T \lambda(t)^\top \partial f / \partial \theta \, \mathrm{d}t,
\end{equation}
making it suitable for training models with terminal loss functions.
However, the adjoint sensitivity method encounters numerical challenges, particularly in stiff or chaotic systems where reverse-time integration can amplify floating-point errors. Alternatively, checkpointing techniques \cite{gholami2019anode,zhuang2020adaptive}, store selected intermediate states during the forward pass, allowing localized recomputation during backward propagation. 

\paragraph{Integral Loss Functions.}
Beyond terminal loss functions, NDEs have been extended to incorporate integral loss functions distributed across the entire depth domain $S$. In the context of optimal control \cite{pontryagin2018mathematical}, the integral loss is defined as formulated by \citet{massaroli2020stable}:
\begin{equation}
    \mathcal{L}(\bm{z}(S)) + \int_\mathcal{S} \ell(t, \bm{z}(t)) \, \mathrm{d}t,    
\end{equation}
where the loss combines terminal contributions \(\mathcal{L}(\bm{z}(S))\) with intermediate terms \(\ell(t, \bm{z}(t))\). This formulation allows the latent state \(\bm{z}(t)\) to evolve through a continuum of layers, guiding the model output toward the desired trajectory over the entire depth domain $S$. The adjoint dynamics for such integral loss functions are modified to include an additional term, \(-\partial \ell / \partial \bm{z}(t)\), accounting for the distributed loss in the backward gradients \cite{grathwohl2018ffjord,massaroli2020stable,finlay2020train}. This approach enhances the control and flexibility of trajectory-level learning, enabling improved performance in tasks that require supervision over the entire temporal or spatial domain.

\paragraph{Backpropagation through time.} 
For discrete methods like RNNs, backpropagation is a common optimization strategy, where the model's recurrence is unfolded into a sequence of differentiable operations. However, these discrete methods face vanishing or exploding gradients over long sequences \cite{hochreiter1997long,pascanu2013difficulty,goodfellow2016deep}. 
NDE-based variants, such as GRU-ODE \cite{de2019gru}, ODE-RNN \cite{rubanova2019latent}, ODE-LSTM \cite{lechner2020learning}, mitigate these issues by replacing discrete recurrence with continuous evolution, enabling robust modeling of time series while retaining mechanisms like gating for temporal dependencies.

\subsection{Regularization Methods}
\paragraph{Advanced Techniques for NDEs.}
While conventional regularization techniques are widely applicable to NDE-based methods, recent advances have introduced specialized approaches tailored to continuous dynamics. Regularization based on principles of optimal transport, proposed by \citet{finlay2020train,finlay2020learning}, simplifies the dynamics of continuous normalizing flow models, thereby accelerating training. \citet{kelly2020learning} analyzed equations with differential surrogates for the computational cost of standard solvers, leveraging higher-order derivatives of solution trajectories to reduce complexity. 

\paragraph{Continuous-Time Modifications.}
Standard regularization techniques have further refined NDE-based methods; stochastic sampling of end times~\cite{ghosh2020steer}, continuous-time dropout~\cite{liu2020does}, and temporal adaptive batch normalization~\cite{zheng2024improving}. These methods form a comprehensive toolkit for advancing the performance and scalability of NDE-based models.

\subsection{Numerical Solution of NDEs}
\paragraph{Comparison with Discrete Architectures.}
Discrete-time methods, including variants of RNNs and Transformers, rely on layer-wise or self-attention updates without numeric integration.
Although this design simplifies backpropagation and typically scales well, these models often struggle to accommodate irregular sampling unless augmented by masking or gating mechanisms \cite{chung2014empirical}. RNN-based architectures can also experience vanishing or exploding gradients for lengthy sequences unless they incorporate techniques such as gradient clipping or gated units \cite{hochreiter1997long,pascanu2013difficulty}. Transformers replace recurrence with attention mechanisms but depend heavily on positional encodings and expansive parameter sets, which can limit their practicality for extensive time spans with variable sampling \cite{vaswani2017attention,rae2019compressive,wen2023transformers}. 
By contrast, continuous-time NDEs use solvers that naturally adapt to asynchronous events, though solver overhead and potential stiffness require careful solver selection and parameter tuning. 

\paragraph{Fixed or Adaptive Solvers.}
NDE-based models rely on numerical integration to evolve their latent states \(\vz(t)\) from initial conditions to final outputs. A fixed-step solver, such as explicit Euler or a basic Runge--Kutta method, updates \(\vz(t)\) at uniform intervals and is computationally simpler but may face instability if the dynamics change abruptly or if the time step is too large \cite{chen2018neural}. Adaptive-step solvers, such as Dormand--Prince method, refine the step size \(\Delta t\) in response to local error estimates, offering higher accuracy for stiff or highly non-uniform dynamics \cite{kidger2021efficient}; however, this adaptive nature can cause unpredictable run times in large-scale tasks \cite{rackauckas2020universal}. In the stochastic setting of NSDEs, solvers like Euler--Maruyama or Milstein handle the Brownian increments for the diffusion term, balancing efficiency and stability under well-defined noise \cite{tzen2019neural,oh2024stable}.

%-----------------------------------------------------------------------
\section{Comparison of NDE-based Methods}
Table~\ref{tab:methods} summarizes the core formulations, primary tasks, and benchmark datasets associated with various NDE-based methods. These methods can be grouped into three main categories—NODE, NCDE, and NSDE—highlighting their unique characteristics and application domains. 

%%%%%
\begin{table*}[ht!]
\scriptsize\centering
% \caption{Comparison of NDE Methods: Formulations, Applications, and Benchmarks}\label{tab:methods}
\begin{tabular}{P{2.8cm} P{7.6cm} P{3.0cm} P{2.8cm}}
\toprule
\textbf{NODE Methods} & \textbf{State Evolution / Formulation} & \textbf{Applications / Tasks} & \textbf{Benchmarks / Datasets} \\
\midrule

\textbf{Neural ODE} \newline \cite{chen2018neural} &
\(\begin{aligned}
\mathrm{d}\bm{z}(t) = f\bigl(t, \bm{z}(t); \theta_f\bigr) \, \mathrm{d}t
\end{aligned}\) &
Continuous normalizing flow, Image classification & 
Synthetic data, MNIST \\
\midrule

\textbf{Augmented ODE} \newline \cite{dupont2019augmented} &
\(\begin{aligned}
\frac{\mathrm{d}}{\mathrm{d}t} 
\begin{pmatrix} 
\bm{z}(t) \\ 
\bm{a}(t) 
\end{pmatrix}
= f\left(
t, 
\begin{pmatrix}
\bm{z}(t) \\ 
\bm{a}(t) 
\end{pmatrix}; \theta_f\right)
\end{aligned}
\) &
Complex dynamics modeling, Image classification & 
Synthetic data, MNIST, CIFAR-10, SVHN, ImageNet \\
\midrule

\textbf{GRU-ODE} \newline \cite{de2019gru} &
\(\begin{aligned}
\frac{\mathrm{d}\bm{z}(t)}{\mathrm{d}t} = \bigl(1 - \bm{u}(t)\bigr) \circ \bigl(\bm{g}(t) - \bm{z}(t)\bigr)
\end{aligned}\) &
Complex dynamics modeling, Time series forecasting & 
Synthetic data, USHCN, MIMIC-III\\
\midrule

\textbf{ODE$^2$VAE} \newline \cite{yildiz2019ode2vae} &
\(\begin{aligned}
& \vz(t) = (\bm{s}(t), \bm{v}(t)), \quad
\dot{\bm{s}}(t) = \bm{v}(t), \quad
\dot{\bm{v}}(t) = f\bigl(\bm{s}(t), \bm{v}(t)\bigr), \\
% \quad {\text{where } f_{\mathcal{W}} \text{ is a BNN,}} \\
& q_\text{enc}(\bm{z}(0) | \bm{x}_{0:N}) = q_\text{enc}\left(\begin{pmatrix} \bm{s}(0) \\ \bm{v}(0) \end{pmatrix} \; \middle| \; \bm{x}_{0:N} \right) \; \text{(variational approximation)}
\end{aligned}
\) &
High-dimensional sequential data modeling, Motion prediction, Image sequence forecasting &
CMU Motion Capture, MNIST, Bouncing Balls \\
\midrule

\textbf{ODE-RNN} \newline \cite{rubanova2019latent} & 
\(\begin{aligned}
& \bar{\bm{h}}_{t} = \text{ODESolve}\bigl(f, \bm{h}_{t-1}, [t_{t-1}, t_t]\bigr) \\
& \bm{h}_{t} = \text{RNNCell}\bigl(\bar{\bm{h}}_{t}, \bm{x}_{t}\bigr)
\end{aligned}\) &
Continuous-time modeling, Interpolation and extrapolation & 
MuJoCo, PhysioNet mortality, Human Activity \\
\midrule

\textbf{ODE-LSTM} \newline \cite{lechner2020learning} &
\(\begin{aligned}
& \bar{\bm{h}}_{t} = \text{ODESolve}\bigl(f, \bm{h}_{t-1}, [t_{t-1}, t_t]\bigr) \\
& \bm{h}_{t}, \bm{c}_{t} = \text{LSTMCell}\bigl(\bm{c}_{t-1}, \bar{\bm{h}}_{t}, \bm{x}_{t}\bigr), \quad \text{where internal memory}\; \bm{c}_t \\ 
\end{aligned}\) &
Continuous-time modeling, Time series classification & 
Synthetic data, Activity recognition, MNIST, MuJoCo physics \\
\bottomrule
\end{tabular}
\vfil\vspace{0.5em}
\begin{tabular}{P{2.8cm} P{7.6cm} P{3.0cm} P{2.8cm}}
\toprule
\textbf{NCDE Methods} & \textbf{State Evolution / Formulation} & \textbf{Applications / Tasks} & \textbf{Benchmarks / Datasets} \\
\midrule

\textbf{Neural CDE} \newline \cite{kidger2020neural} &
\(\begin{aligned}
& \mathrm{d}\bm{z}(t) = f\bigl(\bm{z}(t);\theta_f\bigr)\,\mathrm{d}X(t), \\ 
& \text{using Riemann–Stieltjes integral over control path} \; X(t) 
\end{aligned}\) &
Irregular time series analysis, Time series classification & 
CharacterTrajectories, Speech Commands, PhysioNet Sepsis \\
\midrule

\textbf{Neural RDE} \newline \cite{morrill2021neural} &
\(\begin{aligned}
& \mathrm{d}\bm{z}(t) = f\bigl(\bm{z}(t); \theta_f\bigr) \,\mathrm{d}\logsig_{[r_i, r_{i+1}]}(X), \\
& \text{where} \; \logsig_{[r_i, r_{i+1}]}(X) \; \text{represents the log-signature of} \; X(t) 
\end{aligned}\) &
Long time series analysis, \; Robust feature extraction & 
EigenWorms, BIDMC \\
\midrule
\textbf{Neural Lad} \newline \cite{li2023neural} &
\(
\begin{aligned}
& \frac{dz(t)}{dt}
=
h_{w}(t)\,\phi_{\gamma}\bigl(z(t)\bigr)\,f\bigl(z(t)\bigr)\,
g_{v}\!\Bigl(\tfrac{dX}{dt}\Bigr),\\
\end{aligned}
\) &
Time series forecasting, Irregularly sampled time series classification & 
PhysioNet Sepsis, ETT, Weather, PEMS\\
\midrule

\textbf{ANCDE} \newline \cite{jhin2024attentive} &
\(\begin{aligned}
& \mathrm{d}\bm{h}(t) = f\bigl(\bm{h}(t); \bm{\theta}_f) \, \mathrm{d}X(t) \quad \text{(attention value, \(\bm{a}(t) = \sigma(\bm{h}(t))\))} \\
& \mathrm{d}\bm{z}(t) = g\bigl(\bm{z}(t); \bm{\theta}_g) \, \mathrm{d}Y(t) \quad\; \text{(attention-modulated control path, \(Y(t)\))} 
\end{aligned}\) &
Irregular time series classification and forecasting & 
CharacterTrajectories, PhysioNet Sepsis, Google stock, MuJoCo physics\\
\midrule

\textbf{DualDynamics} \newline \cite{oh2025dualdynamics} &
\(\begin{aligned}
& \frac{\mathrm{d}\bm{z}(t)}{\mathrm{d}t} = f\bigl(\bm{z}(t);\theta_f\bigr)\,\mathrm{d}X(t) := f^{*}(t; \theta_{f}, X) \, \rd t \\ 
& \hat{\vz}(t) = \gG\bigl(t, \hat{\vz}(0); \theta_\gG\bigr) \quad \text{where}\; \frac{\mathrm{d} \gG\bigl(t,\hat{\vz}(0)\bigr)}{\mathrm{d} t} =f^*(t;\theta_f,X)
\end{aligned}\) &
Irregular time series classification, imputation, and forecasting & 
UEA/UCR archive, PhysioNet mortality, PhysioNet Sepsis, Google stock, MuJoCo physics \\ 
\bottomrule
\end{tabular}
\vfil\vspace{0.5em}
\begin{tabular}{P{2.8cm} P{7.6cm} P{3.0cm} P{2.8cm}}
\toprule
\textbf{NSDE Methods} & \textbf{State Evolution / Formulation} & \textbf{Applications / Tasks} & \textbf{Benchmarks / Datasets} \\
\midrule

\textbf{Neural SDE} \newline \cite{tzen2019neural} &
\(\begin{aligned}
\mathrm{d}\bm{z}(t) = f\bigl(t, \vz(t); \theta_f\bigr) \, \mathrm{d}t + g\bigl(t, \vz(t); \theta_g\bigr) \, \mathrm{d}W(t)
\end{aligned}\) &
Stochastic dynamics modeling with variational inference framework & 
Theoretical analysis \\
\midrule

\textbf{Neural Jump SDE} \newline \cite{jia2019neural} &
\(\begin{aligned}
\mathrm{d}\bm{z}(t) = f\bigl(t, \vz(t); \theta_f\bigr) \, \mathrm{d}t + w(t, \vz(t), \vk(t); \theta_w) \cdot \mathrm{d}N(t)
\end{aligned}\) &
Point process modeling, Event feature prediction & 
Synthetic data, Stack Overflow, MIMIC-II, Earthquake\\
\midrule

\textbf{Latent SDE} \newline \cite{li2020scalable} &
\(\begin{aligned}
& \mathrm{d}\widetilde{\bm{z}}(t) = h_\theta\bigl(t, \widetilde{\bm{z}}(t)\bigr) \, \mathrm{d}t + \sigma\bigl(t, \widetilde{\bm{z}}(t)\bigr) \, \mathrm{d}W(t) \quad \; \text{(prior)} \\ 
& \mathrm{d}\bm{z}(t) = h_\phi\bigl(t, \bm{z}(t)\bigr) \, \mathrm{d}t + \sigma\bigl(t, \bm{z}(t)\bigr) \, \mathrm{d}W(t) \quad \; \text{(approximate posterior)} \\ 
\end{aligned}\) &
Stochastic adjoint sensitivity method & 
Synthetic data, Motion capture data \\
\midrule

\textbf{Neural SDEs as GANs} \newline \cite{kidger2021neuralsde} &
\(\begin{aligned}
& \mathrm{d}\vx(t) = \mu_\theta(t, \vx(t)) \, \mathrm{d}t + \sigma_\theta(t, \vx(t)) \, \mathrm{d}W(t) \quad \text{(generator)} \\
& \mathrm{d}\vh(t) = f_\phi(t, \vh(t)) \, \mathrm{d}t + g_\phi(t, \vh(t)) \, \mathrm{d}Y(t) \quad\; \text{(discriminator)}
\end{aligned}\) &
Generative adversarial network, Time series classification and prediction & 
Synthetic data, Google/Alphabet stock,  Beijing air quality\\
\midrule

\textbf{Stable Neural SDEs} \newline \cite{oh2024stable} &
\(\begin{aligned}
& \textbf{Neural LSDE:} &    
 \rd\vz(t) &= \gamma\bigl(\overline{\vz}(t);\theta_\gamma\bigr) \, \rd t + \sigma(t;\theta_\sigma) \, \rd W(t) \\[0.2em] 
& \textbf{Neural LNSDE:} &  
 \rd\vz(t) &= \gamma\bigl(t, \overline{\vz}(t);\theta_\gamma\bigr) \, \rd t + \sigma(t;\theta_\sigma)\vz(t) \, \rd W(t) \\[0.2em]
& \textbf{Neural GSDE:} &   
 \frac{\rd\vz(t)}{\vz(t)} &=  \gamma\bigl(t,\overline{\vz}(t);\theta_\gamma\bigr) \, \rd t + \sigma(t;\theta_\sigma) \, \rd W(t)
\end{aligned}\) 
&
Irregular time series classification, imputation, and forecasting & 
UEA/UCR archive, Speech Commands, PhysioNet mortality, PhysioNet Sepsis, MuJoCo physics \\ 
\bottomrule
\end{tabular}
\caption{Comparison of NDE Methods: Formulations, Applications, and Benchmarks}\label{tab:methods}
\end{table*}
%%%%%

\subsection{NODE Methods}
Standard form of Neural ODE \cite{chen2018neural} introduced a novel approach to continuous-time modeling by parameterizing the hidden state dynamics with a neural network, allowing for the seamless handling of irregularly sampled time series data. 
Augmented ODE \cite{dupont2019augmented} extend this framework by augmenting the state space with auxiliary variables, enabling the modeling of complex trajectories and avoiding trajectory overlap. %
ODE$^2$VAE~\cite{yildiz2019ode2vae} combines NODEs using Bayesian Neural Network (BNN) with variational autoencoders to model latent dynamics in irregular time series by learning both the continuous-time evolution and the underlying latent space.

On the other hand, GRU-ODE \cite{de2019gru} adopts continuous gating mechanisms inspired by GRU cells, where the hidden state evolves dynamically based on time gaps. $\bm{u}(\cdot)$, and $\bm{g}(\cdot)$ are continuous counterparts of GRU. 
ODE-RNN \cite{rubanova2019latent} combines ODE-based continuous evolution with discrete updates at observation points using an RNN cell. 
Similar to that, ODE-LSTM \cite{lechner2020learning} incorporates the memory mechanisms of LSTM into the ODE framework, evolving the hidden state between observations using ODE solvers and performing discrete LSTM updates at observation points.
%
% While these methods enhance NODEs, they still rely on the initial conditions.

\subsection{NCDE Methods}
Neural CDE~\cite{kidger2020neural} models the latent state \(\bm{z}(t)\) with a piecewise-smooth control path \(X(t)\). The Riemann–Stieltjes integral allows Neural CDEs to handle irregular or asynchronous time series data effectively, making them suitable for tasks like interpolation and forecasting.
Neural Rough Differential Equations (Neural RDEs)~\cite{morrill2021neural} extend Neural CDEs by using the log-signature of the control path \(\logsig_{[r_i, r_{i+1}]}(X)\) with \(X:[t_0:t_n]\) and certain intervals \(t_0 \le r_{i} < r_{i+1}\le t_n\), capturing higher-order variations. This enables Neural RDEs to handle rough paths, improving robustness for long and complex time series. Recently, \citet{walker2024log} further extended rough path theory to Log-NCDEs.
These approaches leverage continuous-time dynamics, enabling flexible modeling of time series.

Recent works \cite{jhin2022exit,jhin2023learnable} emphasize training-based approaches for constructing control paths in NCDEs. For instance, Attentive Neural Controlled Differential Equation (ANCDE)~\cite{jhin2024attentive} extend this line of research by incorporating attention mechanisms for dynamic path construction. This is implemented using two coupled NCDEs: one computes attention values, \(\bm{a}(t) = \sigma(\bm{h}(t))\) using an activation \(\sigma\), and the other applies the attention-modulated control path, \(Y(t)\), which is a combination of vector \(a(t)\) and \(X(t)\).
Similarly, \citet{li2023neural} propose Neural LAD (Latent dynamics), which incorporates an attention-based transformation of the input path gradient
\(g_{v} \tfrac{dX}{dt}\) for continuous-time latent dynamics. In addition, Neural LAD employs a seasonality-trend transformation $h_w{(t)}$ to capture periodic and slowly varying patterns, and uses \(\phi_{\gamma}\bigl(\bm{z}(t)\bigr)\) to learn structural relationships within the latent state. % By unifying these components, Neural Lad effectively handles irregular sampling, long-horizon forecasting, and missing observations, making it suitable for a wide range of real-world time series tasks.

DualDynamics~ \cite{oh2025dualdynamics} combines explicit and implicit mechanisms within a unified framework to improve the modeling of irregular time series data. By integrating Neural ODEs for explicit time evolution and learnable implicit updates for latent state transitions, DualDynamics achieves a balance between interpretability and robustness. 
These variations of NCDEs have been extensively applied to irregularly-sampled time series or data with missing observations, demonstrating their effectiveness in a variety of tasks including classification, imputation, and forecasting.

\subsection{NSDE Methods}
Neural SDEs, parameterized by neural networks for both drift and diffusion terms, provide a flexible framework for modeling complex stochastic processes, with foundational contributions from \citet{han2017deep} and \citet{tzen2019neural}.
Neural Jump SDE~\cite{jia2019neural} incorporates Poisson jump processes to model discontinuous state changes. \citet{zhang2024neural} extended this to temporal point processes. While requiring specialized solvers, this approach effectively captures rare, discontinuous events in NSDEs.

Latent SDE~\cite{li2020scalable} extends the encoding-decoding paradigm of Latent ODE~\cite{rubanova2019latent} by incorporating stochastic dynamics, allowing for more flexible modeling of uncertainty and variability. Latent SDE defines a prior process \(\widetilde{\bm{z}}(t)\) and an approximate posterior process \(\bm{z}(t)\), both parameterized as Neural SDEs with \(h_\theta\), \(h_\phi\), and \(\sigma\). 
During the training, the Kullback–Leibler (KL) divergence measures and minimizes the prior-posterior discrepancy while maintaining uncertainty.
On the other hand, \citet{kidger2021neuralsde} extends Wasserstein generative adversarial network (GAN) to continuous-time stochastic processes, where both generator and discriminator are formulated as Neural SDEs.

% \citet{oh2024stable} introduced Stable Neural SDEs with three variants: 
Stable Neural SDEs, introduced in \citet{oh2024stable}, extend Neural SDEs to address stability and robustness in complex systems: 
Neural Langevin-type SDE (LSDE) with additive noise for invariant measures, Neural Linear Noise SDE (LNSDE) with multiplicative noise for scale-dependent fluctuations, and Neural Geometric SDE (GSDE) with state-dependent noise for exponential dynamics. 
Note that, these formulations utilize the augmented states with control path, \(\overline{\vz}(t) = \zeta(t, \vz(t), X(t); \theta_\zeta)\), in order to ensure stability for irregular time series analysis.
Therefore, NSDEs extend NODEs by incorporating stochasticity, enabling robust modeling of complex, uncertain, and discontinuous dynamics.

\subsection{Empirical Comparison}
We present a consolidated summary of diverse empirical results from \citet{oh2024stable} and \citet{oh2025dualdynamics}, encompassing tasks such as interpolation on PhysioNet Mortality, forecasting on MuJoCo, and classification on PhysioNet Sepsis.

We summarize interpolation experiments on the 2012 PhysioNet Mortality dataset~\cite{silva2012predicting}, which contains 37 ICU variables recorded over 48 hours. Following~\citet{rubanova2019latent}, timestamps were rounded to the nearest minute, resulting in up to 2880 intervals. As in~\citet{shukla2021multi}, observation rates varied from 50\% to 90\%, using 8000 instances to predict the unobserved values.
Benchmark models included RNN-VAE~\cite{chen2018neural}, 
mTAND-Full~\cite{shukla2021multi}, 
% TimeCHEAT~\cite{liu2025timecheat},
L-ODE-RNN~\cite{chen2018neural}, 
L-ODE-ODE~\cite{rubanova2019latent},
DualDynamics~\cite{oh2025dualdynamics},
Latent SDE~\cite{li2020scalable}, 
and variants of Neural SDEs~\cite{oh2024stable}. 
Well-designed NDE methods outperform conventional models in interpolation performance, as shown in Table~\ref{tab:interpolation}.
%%%
\begin{table}[htb]
\scriptsize\centering\captionsetup{skip=5pt}
% \caption{Interpolation performance (Test MSE $\times 10^{-3}$) versus percent observed time points on PhysioNet Mortality dataset}\label{tab:interpolation}
\begin{tabular}{@{}m{1.8cm}C{0.8cm}C{0.8cm}C{0.8cm}C{0.8cm}C{0.8cm}@{}}
\toprule
\multicolumn{1}{c}{\multirow{2.5}{*}{\textbf{Methods}}} & \multicolumn{5}{c}{\textbf{Observed \%}}                                           \\ \cmidrule(l){2-6} 
\multicolumn{1}{c}{}                                  & \textbf{50\%}  & \textbf{60\%}  & \textbf{70\%}  & \textbf{80\%}  & \textbf{90\%}  \\ \midrule
RNN-VAE                                               & 13.418   \std{0.008} & 12.594 \std{0.004} & 11.887 \std{0.005} & 11.133 \std{0.007} & 11.470 \std{0.006} \\
mTAND-Full                                            & 4.139 \std{0.029}    & 4.018 \std{0.048}  & 4.157 \std{0.053}  & 4.410 \std{0.149}  & 4.798 \std{0.036}  \\
% TimeCHEAT                                             & 4.185 \std{0.030}    & 3.981 \std{0.016}  & 3.657 \std{0.022}  & 3.642 \std{0.036}  & 3.686 \std{0.009}  \\
L-ODE-RNN                                             & 8.132 \std{0.020}    & 8.140 \std{0.018}  & 8.171 \std{0.030}  & 8.143 \std{0.025}  & 8.402 \std{0.022}  \\
L-ODE-ODE                                             & 6.721 \std{0.109}    & 6.816 \std{0.045}  & 6.798 \std{0.143}  & 6.850 \std{0.066}  & 7.142 \std{0.066}  \\
DualDynamics                                          & 3.631 \std{0.049}    & 3.659 \std{0.028}  & 3.463 \std{0.032}  & 3.224 \std{0.044}  & 3.114 \std{0.050}  \\
Latent SDE                                             & 8.862 \std{0.036}    & 8.864 \std{0.058}  & 8.686 \std{0.122}  & 8.716 \std{0.032}  & 8.435 \std{0.077}  \\
Neural SDE                                            & 8.592 \std{0.055}    & 8.591 \std{0.052}  & 8.540 \std{0.051}  & 8.318 \std{0.010}  & 8.252 \std{0.023}  \\
Neural LSDE                                           & 3.799 \std{0.055}    & 3.584 \std{0.055}  & 3.457 \std{0.078}  & 3.262 \std{0.032}  & 3.111 \std{0.076}  \\
Neural LNSDE                                          & 3.808 \std{0.078}    & 3.617 \std{0.129}  & 3.405 \std{0.089}  & 3.269 \std{0.057}  & 3.154 \std{0.084}  \\
Neural GSDE                                           & 3.824 \std{0.088}    & 3.667 \std{0.079}  & 3.493 \std{0.024}  & 3.287 \std{0.070}  & 3.118 \std{0.065}  \\ \bottomrule
\end{tabular}
\caption{Interpolation performance (Test MSE $\times 10^{-3}$) versus percent observed time points on PhysioNet Mortality dataset}\label{tab:interpolation}
\end{table}
%%%

For forecasting and classification task, we consider a variety of NDE-related methods, including
GRU-$\Delta t$~\cite{choi2016doctor}, 
GRU-D~\cite{che2018recurrent},
GRU-ODE~\cite{de2019gru}, 
ODE-RNN~\cite{rubanova2019latent},
% ODE-LSTM~\cite{lechner2020learning}, 
Latent-ODE~\cite{rubanova2019latent}, 
Augmented-ODE~\cite{dupont2019augmented}, 
Attentive co-evolving neural ordinary differential equations (ACE-NODE)~\cite{jhin2021ace},
Neural CDE~\cite{kidger2020neural},
% Neural RDE~\cite{morrill2021neural},
ANCDE~\cite{jhin2024attentive}, 
EXtrapolation and InTerpolation-based model (EXIT)~\cite{jhin2022exit}, 
LEArnable Path-based model (LEAP)~\cite{jhin2023learnable}, 
DualDynamics~\cite{oh2025dualdynamics},
and variants of Neural SDEs~\cite{oh2024stable} including Neural LSDE, Neural LNSDE, and Neural GSDE. 

Table~\ref{tab:forecasting} summarizes forecasting results, which used the MuJoCo dataset~\cite{todorov2012mujoco} 
with the Hopper setup from the DeepMind control suite~\cite{tassa2018deepmind}. Following~\citet{jhin2022exit,jhin2023learnable}, models were trained to forecast 10 future steps from 50 observed points under 30\%, 50\%, and 70\% missingness. 
Similar to the previous results, NDE models demonstrate both superior prediction performance under varying missingness and greater memory efficiency in general, in MuJoCo forecasting.
%%%
\begin{table}[htb]
\scriptsize\centering\captionsetup{skip=5pt}
% \caption{Forecasting MSE and memory usage versus percent dropped time points on MuJoCo  dataset}\label{tab:forecasting}
\begin{tabular}{@{}m{1.8cm}C{0.8cm}C{0.8cm}C{0.8cm}C{0.8cm}C{0.8cm}C{1.0cm}@{}}
\toprule
\multicolumn{1}{c}{\multirow{3.5}{*}{\textbf{Methods}}} & \multicolumn{4}{c}{\textbf{Test MSE}}                                                     & \multirow{3.5}{*}{
\begin{tabular}[c]{@{}c@{}}\textbf{Memory} \\ \textbf{(MB)}\end{tabular}} \\ \cmidrule(lr){2-5}
\multicolumn{1}{c}{}                                  & \textbf{0\% dropped}  & \textbf{30\% dropped} & \textbf{50\% dropped} & \textbf{70\% dropped} &                                  \\ \midrule
GRU-$\Delta t$                                        & 0.223 \std{0.020} & 0.198 \std{0.036}     & 0.193 \std{0.015}     & 0.196 \std{0.028}     & 533                              \\
GRU-D                                                 & 0.578 \std{0.042} & 0.608 \std{0.032}     & 0.587 \std{0.039}     & 0.579 \std{0.052}     & 569                              \\
GRU-ODE                                               & 0.856 \std{0.016} & 0.857 \std{0.015}     & 0.852 \std{0.015}     & 0.861 \std{0.015}     & 146                              \\
ODE-RNN                                               & 0.328 \std{0.225} & 0.274 \std{0.213}     & 0.237 \std{0.110}     & 0.267 \std{0.217}     & 115                              \\
Latent-ODE                                            & 0.029 \std{0.011} & 0.056 \std{0.001}     & 0.055 \std{0.004}     & 0.058 \std{0.003}     & 314                              \\
Augmented-ODE                                         & 0.055 \std{0.004} & 0.056 \std{0.004}     & 0.057 \std{0.005}     & 0.057 \std{0.005}     & 286                              \\
ACE-NODE                                              & 0.039 \std{0.003} & 0.053 \std{0.007}     & 0.053 \std{0.005}     & 0.052 \std{0.006}     & 423                              \\
Neural CDE                                            & 0.028 \std{0.002} & 0.027 \std{0.000}     & 0.027 \std{0.001}     & 0.026 \std{0.001}     & 52.1                             \\
ANCDE                                                 & 0.026 \std{0.001} & 0.025 \std{0.001}     & 0.025 \std{0.001}     & 0.024 \std{0.001}     & 79.2                             \\
EXIT                                                  & 0.026 \std{0.000} & 0.025 \std{0.004}     & 0.026 \std{0.000}     & 0.026 \std{0.001}     & 127                              \\
LEAP                                                  & 0.022 \std{0.002} & 0.022 \std{0.001}     & 0.022 \std{0.002}     & 0.022 \std{0.001}     & 144                              \\
DualDynamics                                          & 0.006 \std{0.001} & 0.008 \std{0.001}     & 0.008 \std{0.001}     & 0.008 \std{0.001}     & 318                              \\
Neural SDE                                            & 0.028 \std{0.004} & 0.029 \std{0.001}     & 0.029 \std{0.001}     & 0.027 \std{0.000}     & 234                              \\
Neural LSDE                                           & 0.013 \std{0.000} & 0.014 \std{0.001}     & 0.013 \std{0.000}     & 0.013 \std{0.001}     & 249                              \\
Neural LNSDE                                          & 0.012 \std{0.001} & 0.014 \std{0.001}     & 0.013 \std{0.001}     & 0.014 \std{0.000}     & 273                              \\
Neural GSDE                                           & 0.013 \std{0.001} & 0.013 \std{0.001}     & 0.013 \std{0.000}     & 0.014 \std{0.000}     & 306                              \\ \bottomrule
\end{tabular}
\caption{Forecasting MSE and memory usage versus percent dropped time points on MuJoCo  dataset}\label{tab:forecasting}
\end{table}
%%%

Table~\ref{tab:classification} summarizes classification results, which used the PhysioNet Sepsis dataset~\cite{reyna2019early}, containing 40,335 patients and 34 temporal variables. Following~\citet{kidger2020neural}, we compared classification with and without observation intensity (OI), an index reflecting patient severity. Classification performance was measured by AUROC.
On the PhysioNet Sepsis task, NDE-based models consistently achieve higher AUROC with significantly lower memory usage compared to conventional baselines.
%%%
\begin{table}[htb]
\scriptsize\centering\captionsetup{skip=5pt}
% \caption{AUROC and memory usage on PhysioNet Sepsis}\label{tab:classification}
\begin{tabular}{@{}lccccc@{}} 
\toprule
\multirow{2.5}{*}{\textbf{Methods}} & \multicolumn{2}{c}{\textbf{Test AUROC}} & & \multicolumn{2}{c}{\textbf{Memory (MB)}} \\ \cmidrule{2-3}\cmidrule{5-6} 
\multicolumn{1}{c}{}  & \multirow{1}{*}{\textbf{OI}}  & \multirow{1}{*}{\textbf{No OI}}  & & \multirow{1}{*}{\textbf{OI}} & \multirow{1}{*}{\textbf{No OI}} \\ \midrule
GRU-$\Delta t$ & 0.878 \std{0.006} & 0.840 \std{0.007} & & 837 & 826  \\
GRU-D & 0.871 \std{0.022} & 0.850 \std{0.013} & & 889 & 878  \\
GRU-ODE & 0.852 \std{0.010} & 0.771 \std{0.024} & & 454 & 273  \\
ODE-RNN & 0.874 \std{0.016} & 0.833 \std{0.020} & & 696 & 686  \\
Latent-ODE & 0.787 \std{0.011} & 0.495 \std{0.002} & & 133 & 126  \\
ACE-NODE & 0.804 \std{0.010} & 0.514 \std{0.003} & & 194 & 218  \\
Neural CDE  & 0.880 \std{0.006} & 0.776 \std{0.009} & & 244 & 122 \\
ANCDE & 0.900 \std{0.002} & 0.823 \std{0.003} & & 285 & 129 \\ 
EXIT & 0.913 \std{0.002} & 0.836 \std{0.003} & & 257 & 127 \\ 
DualDynamics & 0.918 \std{0.003} & 0.873 \std{0.004} & & 453 & 233 \\ 
Neural SDE & 0.799 \std{0.007} & 0.796 \std{0.006} & & 368 & 240 \\ 
Neural LSDE & 0.909 \std{0.004} & 0.879 \std{0.008} & & 373 & 436 \\ 
Neural LNSDE & 0.911 \std{0.002} & 0.881 \std{0.002} & & 341 & 445 \\ 
Neural GSDE & 0.909 \std{0.001} & 0.884 \std{0.002} & & 588 & 280 \\ 
\bottomrule
\end{tabular}
\caption{AUROC and memory usage on PhysioNet Sepsis}\label{tab:classification}
\end{table}
%%%

Across diverse time series related tasks, NDE-based methods consistently achieve better performance compared to conventional models. Their advantages are particularly pronounced in dynamic prediction settings and irregularly sampled data, confirming their suitability for real-world time series modeling in a continuous-time manner.

%-----------------------------------------------------------------------
\section{Discussion and Future Directions}\label{sec:Discussion}
% Having surveyed NDE-based approaches for time series, this section addresses key challenges and future research avenues. We explore computational and theoretical hurdles, alongside promising directions like Physics-Informed NDEs and hybrid architectures vital for advancing the field.

\paragraph{Computational Challenges.}
NDEs face scalability challenges with continuous-time solvers. Fixed-step integrators or adaptive integrators, while accurate, can lead to unpredictable computation times in stiff regions \cite{chen2018neural,kim2021stiff,kidger2021on}. Specialized techniques like parallel-in-time integration and GPU acceleration offer promising solutions \cite{rackauckas2020universal,gholami2019anode}.
Future research should focus on developing comprehensive integration schemes that can automatically balance computational efficiency with numerical accuracy, particularly for large-scale applications in real-time settings.

\paragraph{Theoretical Development.}
Current theoretical understanding requires expansion, particularly for non-stationary and noisy data. While stability guarantees exist for NODEs and NCDEs through Lipschitz constraints \cite{massaroli2020dissecting}, similar guarantees for NSDEs remain incomplete \cite{oh2024stable}. Drift-diffusion analysis under strong noise conditions needs further investigation \cite{tzen2019neural,li2020scalable}.
Development of rigorous frameworks for analyzing convergence properties and error bounds in these stochastic settings would significantly advance both theoretical foundations and practical applications of NDEs.

\paragraph{Physics-Informed NDEs.}
Physics-Informed NDEs (or Physics-Informed Neural Networks) incorporate domain knowledge, such as Partial Differential Equations (PDEs) constraints and conservation laws, to align modeled dynamics with physical systems. 
Recent advances in physics-informed architectures have demonstrated remarkable success in capturing multi-scale phenomena and handling noisy measurements in complex physical systems \cite{rudy2017data,raissi2019physics,cuomo2022scientific}. 
By embedding structural priors, these approaches ensure interpretability and reliability in real-world tasks.

\paragraph{Hybrid Architectures.}
Hybrid architectures combine NDEs with models like graph neural networks or Transformers to capture complex time series patterns. Graph-based NDEs handle spatio-temporal data \cite{poli2019graph,choi2022graph}, while attention mechanisms address long-range dependencies and irregular sampling \cite{jhin2024attentive,li2023neural}. Furthermore, \citet{oh2025dualdynamics} integrates explicit and implicit mechanisms for irregular time series analysis.
Future research directions should focus on developing more efficient training algorithms for these hybrid architectures, and exploring applications in real-world domains where both temporal dynamics and structural relationships are important.

%-----------------------------------------------------------------------
\section{Conclusion}\label{sec:Conclusion}
This paper has provided a focused and comprehensive survey of NDEs, specifically for time series analysis. While existing literature includes reviews on broader topics or specific niches within this domain, a systematic consolidation and comparative analysis of the core architectural families and their variants, tailored to the unique challenges of temporal data, has been largely absent. Our work addresses this gap, offering a structured synthesis of advancements and outlining future research directions in this new paradigm.

NDEs represent a significant advancement in time series modeling, offering a principled approach to handling continuous dynamics. Through various formulations, including NODEs, NCDEs, and NSDEs, they provide flexible frameworks for complex temporal data, including cases with irregular sampling or missing data.
While challenges remain in computational efficiency and theoretical understanding, ongoing developments in solver techniques, stability analysis, and hybrid architectures continue to enhance their capabilities. 
Beyond current applications in classification, interpolation, and forecasting, NDEs show potential in anomaly detection, reinforcement learning, and multi-agent systems. As the field evolves, NDEs are increasingly positioned to address sophisticated time series challenges across diverse applications. 

\section*{Acknowledgments}
\justify{
This research was supported by 
the Basic Science Research Program through the National Research Foundation of Korea (NRF) funded by the Ministry of Education (RS-2024-00407852);
the Korea Health Technology R\&D Project through the Korea Health Industry Development Institute (KHIDI), funded by the Ministry of Health and Welfare, Republic of Korea (HI19C1095);
the National Research Foundation of Korea (NRF) grant funded by the Korea government (MSIT) (RS-2025-00563597);
the Ministry of Trade, Industry and Energy (MOTIE) and Korea Institute for Advancement of Technology (KIAT) through the International Cooperative R\&D program (No.P0025828);
and the Institute of Information \& Communications Technology Planning \& Evaluation (IITP) grant funded by the Korea government (MSIT) (RS-2024-00443780, Development of Foundation Models for Bioelectrical Signal Data and Validation of Their Clinical Applications: A Noise-and-Variability Robust, Generalizable Self-Supervised Learning Approach).
}

%% The file named.bst is a bibliography style file for BibTeX 0.99c
{\small\setstretch{1.0}
\bibliographystyle{named}
\bibliography{references}
}

\end{document}